# A Unified Robust Motion Controller Synthesis for Compliant Robots Driven by Series Elastic Actuators

Emre Sariyildiz
School of Mechanical, Materials, Mechatronic and Biomedical Engineering
Faculty of Engineering and Information Sciences, University of Wollongong,
Northfields Avenue Wollongong NSW 2522 Australia
emre@uow.edu.au

*Abstract*— **This paper proposes a unified robust motion controller for the position and force control problems of compliant robot manipulators driven by Series Elastic Actuators (SEAs). It is shown that the dynamic model of the compliant robot includes not only matched but also mismatched disturbances that act on the system through a different channel from the control input. To tackle this complex robust control problem, the unified robust motion controller is synthesised by employing a second-order Disturbance Observer (DOb), which allows us to estimate not only disturbances but also their first and second order derivatives, and a novel controller design approach in state space. By using the Brunovsky canonical form transformation and the estimations of disturbances and their first and second order derivatives, the dynamic model of the robot is reconstructed so that a new system model that includes only matched disturbances is obtained for compliant robots driven by SEAs. The robust position and force controllers are simply designed by eliminating the matched disturbances of the reconstructed system model via the conventional DOb-based robust control method. The stability and performance of the proposed robust motion controllers are verified by simulations.**

## I. Introduction

Compared to traditional industrial robots that can repeat high-precision position control tasks in a strictly restricted and constructed workspace at factories, next generation robots, such as cobots, exoskeletons, and humanoids, are expected to perform physical interaction control tasks alongside human beings in an open environment [1–5]. When it comes to interacting with an unknown and dynamic environment, safety and high-performance force control become essential requirements for a robotic system [1, 5]. To meet these requirements, compliant and soft robots have received increasing attention over the past decades [6 – 8].

Several compliant and soft actuators have been proposed to improve safety and force control performance in physical robot-environment interaction [6 – 9]. Among them, SEAs are one of the most popular compliant actuators in developing intrinsically safe compliant robotic systems. From orthoses and prostheses to humanoids, quadrupeds, and cobots, many advanced robotic systems have been developed using SEAs in the last two decades [11–14]. An SEA is designed by placing a compliant mechanical component, such as a torsional spring, between the motor and link of the actuator [10]. This compliant mechanical component provides several superiorities over conventional rigid actuators when conducting robot-environment interaction tasks, including safety, low-impact force at motor and gearhead, low-cost force measurement, and high backdrivability [10]. However, the motion control problem, particularly the position control problem, of an SEA is more complicated than that of conventional rigid actuators [6, 11]. This notably limits the practical applications of compliant robots driven by SEAs.

To improve the stability and performance of interaction tasks, many different force/impedance controllers have been proposed for SEAs in the last two decades [15 – 18]. It is shown that internal disturbances, such as friction and hysteresis, have a significant impact on force control when conventional PID controllers are applied to SEAs [15, 18]. To tackle the severe sensitivity problem of SEAs, different robust force controllers have been proposed in the literature. A recent benchmarking study shows that the robust force controllers can significantly improve the stability and performance of SEAs in practice [19].

Despite several important applications such as foot clearance and centre-of-mass adjustment of a compliant humanoid, there are less studies on the position control problem of SEAs [20]. Position control of an SEA is a challenging problem because the link of the actuator is very sensitive to internal and external disturbances [21, 22]. The conventional position controllers proposed for the compensation of compliance error, e.g., conventional PID controller [23], Singular perturbation-based controller [24], and μ-synthesis-based robust controller [25], can provide limited performance in real implementations of SEAs. To tackle this problem, Sariyildiz et al. has recently developed robust position controllers for SEAs [21, 22]. However, the position controllers have yet to be applied to compliant robots.

To this end, this paper proposes robust position and force controllers for compliant robots driven by SEAs in a unified framework. First, a simple yet practical decentralised dynamic model is derived for compliant robots in state space. This model shows that matched and mismatched disturbances have a notable impact on the system. Second, a unified robust motion controller is proposed for the position and force control problems of compliant robots by combining a second-order DOb with a novel performance controller in state space. While the second-order DOb estimates the disturbances and their first and second order derivatives, the state space controller allows one to adjust the position and force control performance by suppressing the matched and mismatched disturbances in practice. Last, the proposed robust motion controller is, for the first time, applied to a compliant robot manipulator driven by SEAs, and its stability and performance are verified by simulations.

The rest of the paper is organised as follows. The dynamic model of a compliant robot driven by SEAs is presented in Section II. The robust position and force controllers are proposed in a unified framework in Section III. Simulation results are presented in Section IV. The paper is concluded in Section V.

## II. DYNAMIC MODEL OF A COMPLIANT ROBOT PERFORMING FREE AND CONTACT MOTIONS

This section derives a simple yet practical dynamic model for a compliant robot driven by SEAs. This section also describes the robust position and force control problems of compliant robots.

### A. Dynamic Model

Figure 1 illustrates the dynamic model of the $i^{th}$ link of the compliant robot in free and contact motions. In this figure, the following apply.

| | |
|---|---|
| $J$ and $m$ | inertia/mass of motor and link; |
| $b_J$ and $b_m$ | friction coefficients at motor and link; |
| $q, \dot{q}$ and $\ddot{q}$ | angle/position, velocity, and acceleration; |
| $\tau_J$ and $\tau_J^d$ | thrust and disturbance torques of motor; |
| $k$ | stiffness of the SEA's spring; |
| $\tau_m^a$ | applied external torque at link; |
| $M, D$ and $K$ | mass, damping, and stiffness (environment); |
| $\bullet_J, \bullet_m$ and $\bullet_e$ | parameters of motor, link, and environment; |
| $\bullet_n$ | nominal parameter of $\bullet$; and |
| $\bullet_i$ | parameter of $\bullet$ at the $i^{th}$ joint. |

By using Fig. 1, the dynamic model of a compliant robot can be derived as follows:

$$\mathbf{m}(\mathbf{q_m})\ddot{\mathbf{q}}_\mathbf{m} + \mathbf{c}(\mathbf{q_m},\dot{\mathbf{q}}_\mathbf{m})\dot{\mathbf{q}}_\mathbf{m} + \mathbf{b}_\mathbf{m}\dot{\mathbf{q}}_\mathbf{m} + \mathbf{g}(\mathbf{q_m}) = \mathbf{k}(\mathbf{q_J}-\mathbf{q_m}) - \boldsymbol{\tau}_\mathbf{m}^\mathbf{d}$$
$$\mathbf{J}\ddot{\mathbf{q}}_\mathbf{J} + \mathbf{b}_\mathbf{J}\dot{\mathbf{q}}_\mathbf{J} = \boldsymbol{\tau}_\mathbf{J} - \mathbf{k}(\mathbf{q_J}-\mathbf{q_m}) - \boldsymbol{\tau}_\mathbf{J}^\mathbf{d} \quad (1)$$

where $\mathbf{q_J} = [q_{J_1} \cdots q_{J_n}]^T \in \mathbb{R}^n$ and $\mathbf{q_m} = [q_{m_1} \cdots q_{m_n}]^T \in \mathbb{R}^n$ are angle/position vectors of motor and link, respectively; $\dot{\mathbf{q}}_\bullet$ and $\ddot{\mathbf{q}}_\bullet$ are velocity and acceleration vectors, respectively; $\mathbf{m}(\mathbf{q}) \in \mathbb{R}^{n \times n}$ is the inertia matrix of link and $\mathbf{J} \in \mathbb{R}^{n \times n}$ is the diagonal inertia matrix of motor; $\mathbf{c}(\mathbf{q},\dot{\mathbf{q}}) \in \mathbb{R}^{n \times n}$ is the Coriolis and centripetal matrix; $\mathbf{g}(\mathbf{q}) \in \mathbb{R}^n$ is the gravity vector; $\mathbf{b}_\bullet \in \mathbb{R}^{n \times n}$ is the diagonal viscosity matrix; $\mathbf{k} \in \mathbb{R}^{n \times n}$ is the diagonal stiffness matrix; and $\boldsymbol{\tau}_\mathbf{J} \in \mathbb{R}^n$ and $\boldsymbol{\tau}_\bullet^\mathbf{d} \in \mathbb{R}^n$ are the thrust and disturbance torque vectors, respectively [26].

Let us first consider free motion illustrated in Fig. 1a. When the $i^{th}$ link of the actuator does not physically interact with an environment, the dynamic model is derived as follows.

$$J_{n_i}\ddot{q}_{J_i} + b_{J_{ni}}\dot{q}_{J_i} = \tau_{J_i} - k_{n_i}(q_{J_i}-q_{m_i}) - \tau_{J_i}^{dis}$$
$$m_{n_i}\ddot{q}_{m_i} + b_{m_{ni}}\dot{q}_{m_i} = k_{n_i}(q_{J_i}-q_{m_i}) - \tau_{m_i}^{dis} \quad (2)$$

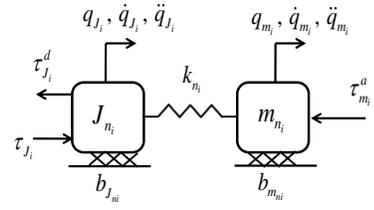

a) Free Motion

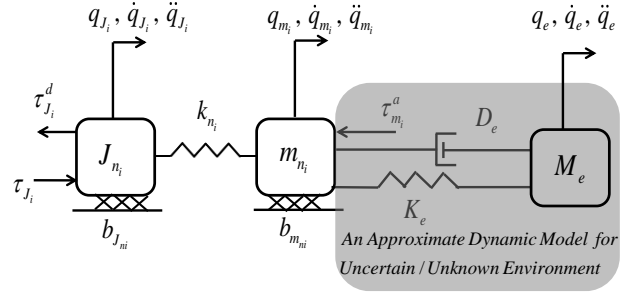

b) Contact Motion.
Fig.1: Dynamic model of an SEA.

where $\tau_{J_i}^{dis}$ and $\tau_{m_i}^{dis}$ are the motor and link disturbances of the $i^{th}$ joint, respectively. Although it is generally impractical to identify these disturbances, they can be simply described using Eq. (3) and Eq. (4).

$$\tau_{J_i}^{dis} = (J_i - J_{n_i})\ddot{q}_{J_i} + (b_{J_i} - b_{J_{ni}})\dot{q}_{J_i} + (k_i - k_{n_i})(q_{J_i} - q_{m_i}) + \tau_{J_i}^u \quad (3)$$

$$\tau_{m_i}^{dis} = (m_i - m_{n_i})\ddot{q}_{m_i} + (b_{m_i} - b_{m_{ni}})\dot{q}_{m_i} + (k_i - k_{n_i})(q_{J_i} - q_{m_i}) + \\ + m_i(\mathbf{q},\ddot{\mathbf{q}}) + c_i(\mathbf{q},\dot{\mathbf{q}}) + g_i(\mathbf{q}) + \tau_{m_i}^a + \tau_{m_i}^u \quad (4)$$

where $m_i(\mathbf{q},\ddot{\mathbf{q}})$, $c_i(\mathbf{q},\dot{\mathbf{q}})$ and $g_i(\mathbf{q})$ represent the interactive disturbance forces due to inertia, Coriolis/centripetal and gravity forces at link i, respectively; and $\tau_{J_i}^u$ and $\tau_{m_i}^u$ represent the unknown/unmodelled disturbances at the motor and link sides, respectively.

Let us now consider the contact motion illustrated in Fig. 1b. The dynamic model of the $i^{th}$ joint can be similarly derived using Eq. (5).

$$J_{n_i}\ddot{q}_{J_i} + b_{J_{ni}}\dot{q}_{J_i} = \tau_{J_i} - k_{n_i}(q_{J_i}-q_{m_i}) - \tilde{\tau}_{J_i}^{dis}$$
$$m_{n_i}\ddot{q}_{m_i} + b_{m_{ni}}\dot{q}_{m_i} = k_{n_i}(q_{J_i}-q_{m_i}) - \tilde{\tau}_{m_i}^{dis} \quad (5)$$

where $\tilde{\tau}_{J_i}^{dis}$ is similar to Eq. (3) but they may vary in practice due to unknown disturbances in free and contact motions, and the disturbance at link side is as follows.

$$\tilde{\tau}_{m_i}^{dis} = (m_i - m_{n_i})\ddot{q}_{m_i} + (b_{m_i} - b_{m_{ni}})\dot{q}_{m_i} + (k_i - k_{n_i})(q_{J_i} - q_{m_i}) + \\ m_i(\mathbf{q},\ddot{\mathbf{q}}) + c_i(\mathbf{q},\dot{\mathbf{q}}) + g_i(\mathbf{q}) + \tau_{m_i}^a + \tau_{m_i}^u + M_e(\ddot{q}_{m_i} - \ddot{q}_e) + \\ D_e(\dot{q}_{m_i} - \dot{q}_e) + K_e(q_{m_i} - q_e) \quad (6)$$

A unified dynamic model can be derived using Eq. (2) and Eq. (5) as follows.

$$\dot{\mathbf{x}}_\mathbf{i} = \mathbf{A}_\mathbf{i}\mathbf{x}_\mathbf{i} + \mathbf{b}_\mathbf{i}u_\mathbf{i} - \boldsymbol{\tau}_{\mathbf{dis}_\mathbf{i}} \quad (7)$$

where $\mathbf{A_i} = \begin{bmatrix} 0 & 1 & 0 & 0 \\ -\frac{k_{n_i}}{J_{n_i}} & -\frac{b_{J_{ni}}}{J_{n_i}} & \frac{k_{n_i}}{J_{n_i}} & 0 \\ 0 & 0 & 0 & 1 \\ \frac{k_{n_i}}{m_{n_i}} & 0 & -\frac{k_{n_i}}{m_{n_i}} & -\frac{b_{m_{ni}}}{m_{n_i}} \end{bmatrix}$, $\mathbf{b_i} = \begin{bmatrix} 0 \\ \frac{1}{J_{n_i}} \\ 0 \\ 0 \end{bmatrix}$, $\boldsymbol{\tau}_{\mathbf{dis_i}} = \begin{bmatrix} 0 \\ \frac{\tau_{dis}^{mtc}}{J_{n_i}} \\ 0 \\ \frac{\tau_{dis}^{mmtc}}{m_{n_i}} \end{bmatrix}$,

$\mathbf{x_i} = \begin{bmatrix} q_{J_i} & \dot{q}_{J_i} & q_{m_i} & \dot{q}_{m_i} \end{bmatrix}^T$, $u = \tau_{J_i}$, $\tau_{dis}^{mtc} = \tau_{J_i}^{dis} \left( \tau_{dis}^{mtc} = \tilde{\tau}_{J_i}^{dis} \right)$ and $\tau_{dis}^{mmtc} = \tau_{m_i}^{dis} \left( \tau_{dis}^{mmtc} = \tilde{\tau}_{m_i}^{dis} \right)$ are the matched and mismatched disturbances in free (contact) motion, respectively.

Let us now describe the decentralised position and force control problems of the compliant robots driven by SEAs.

### B. Robust Position Control Problem

In the robust position control problem, we aim to precisely follow the desired link trajectory by supressing the matched and mismatched disturbances described in Eq. (7). In other words, the robust position controller's performance goal is $q_{m_i} = q_{m_i}^{des}$ where $q_{m_i}^{des}$ represents the desired trajectory of the $i^{th}$ link.

### C. Robust Force Control Problem

By using Hooks law, the robust force control problem is described at the $i^{th}$ link as follows.

$$\tau_{spring} = k_{n_i} \left( q_{J_i} - q_{m_i} \right) = \tau_{spring}^{des} \tag{8}$$

where $\bullet^{des}$ represents the desired $\bullet$.

Equation (8) can also be described as a position control problem by using a simple mathematical manipulation $q_{J_i} - q_{m_i} = \tau_{spring}^{des}/k_{n_i}$, i.e., controlling the relative motion between motor and link. This is one of the most important advantages of SEAs in force control.

### III. ROBUST MOTION CONTROLLER SYNTHESIS

To conduct high-performance robust position and force control applications, the matched and mismatched disturbances given in Eq. (7) should be precisely suppressed. The matched disturbances can be simply eliminated by feeding back the estimated disturbances via the conventional DOb-based robust control method [27]. This conventional method, however, falls-short in tackling mismatched disturbances because there is no control input to directly cancel the effect of such disturbances as shown in Eq. (7). Therefore, a novel robust motion controller is proposed for the compliant robots driven by SEAs in this section.

It is noted that we have recently proposed DOb-based robust motion controllers to precisely suppress the matched and mismatched disturbances of an SEA due to friction, backlash, and external load [6, 11]. However, the interactive disturbances of a compliant robot, e.g., gravity and Coriolis forces, have yet to be considered. By introducing a decentralised robust motion control problem, this paper shows that the proposed robust motion controller can also be similarly applied to compliant robots driven by SEAs.

The decentralised robust motion controller is synthesised by employing a second order DOb and Brunovsky canonical form transformation in state space. Let us start with introducing the second order DOb. The reader is referred to [28] for details of a higher order DOb design.

### A. Second Order DOb

Let us assume that the norm of the disturbance vector and its successive derivatives are bounded, i.e., $\|\boldsymbol{\tau}_{\mathbf{dis_i}}\| \leq \lambda_1$, $\|\dot{\boldsymbol{\tau}}_{\mathbf{dis_i}}\| \leq \lambda_2$, $\|\ddot{\boldsymbol{\tau}}_{\mathbf{dis_i}}\| \leq \lambda_3$ and $\|\dddot{\boldsymbol{\tau}}_{\mathbf{dis_i}}\| \leq \lambda_4$ where $\lambda_\bullet > 0 \in \mathbb{R}$. Let us also describe auxiliary variable vectors $\mathbf{z}_\bullet$ using the state vector $\mathbf{x_i}$, the $i^{th}$ link's disturbance vector $\boldsymbol{\tau}_{\mathbf{dis_i}}$, and its successive derivatives.

$$\begin{bmatrix} \mathbf{z}_{\mathbf{i_1}} \\ \mathbf{z}_{\mathbf{i_2}} \\ \mathbf{z}_{\mathbf{i_3}} \end{bmatrix} = \begin{bmatrix} \mathbf{I} & \mathbf{0} & \mathbf{0} \\ \mathbf{0} & \mathbf{I} & \mathbf{0} \\ \mathbf{0} & \mathbf{0} & \mathbf{I} \end{bmatrix} \begin{bmatrix} \boldsymbol{\tau}_{\mathbf{dis_i}} \\ \dot{\boldsymbol{\tau}}_{\mathbf{dis_i}} \\ \ddot{\boldsymbol{\tau}}_{\mathbf{dis_i}} \end{bmatrix} + \begin{bmatrix} \mathbf{I} & \mathbf{0} & \mathbf{0} \\ \mathbf{0} & \mathbf{I} & \mathbf{0} \\ \mathbf{0} & \mathbf{0} & \mathbf{I} \end{bmatrix} \begin{bmatrix} L_1 \mathbf{x_i} \\ L_2 \mathbf{x_i} \\ L_3 \mathbf{x_i} \end{bmatrix} \tag{9}$$

The estimation of disturbance vector and its successive time derivatives can be obtained using the estimation of auxiliary variables as follows.

$$\begin{bmatrix} \hat{\boldsymbol{\tau}}_{\mathbf{dis_i}} \\ \hat{\dot{\boldsymbol{\tau}}}_{\mathbf{dis_i}} \\ \hat{\ddot{\boldsymbol{\tau}}}_{\mathbf{dis_i}} \end{bmatrix} = \begin{bmatrix} \mathbf{I} & \mathbf{0} & \mathbf{0} \\ \mathbf{0} & \mathbf{I} & \mathbf{0} \\ \mathbf{0} & \mathbf{0} & \mathbf{I} \end{bmatrix} \begin{bmatrix} \hat{\mathbf{z}}_{\mathbf{i_1}} \\ \hat{\mathbf{z}}_{\mathbf{i_2}} \\ \hat{\mathbf{z}}_{\mathbf{i_3}} \end{bmatrix} - \begin{bmatrix} \mathbf{I} & \mathbf{0} & \mathbf{0} \\ \mathbf{0} & \mathbf{I} & \mathbf{0} \\ \mathbf{0} & \mathbf{0} & \mathbf{I} \end{bmatrix} \begin{bmatrix} L_1 \mathbf{x_i} \\ L_2 \mathbf{x_i} \\ L_3 \mathbf{x_i} \end{bmatrix} \tag{10}$$

where $\hat{\bullet}$ represents the estimation of $\bullet$, and the estimations of auxiliary variables are derived by integrating the following equation.

$$\frac{d}{dt} \begin{bmatrix} \hat{\mathbf{z}}_{\mathbf{i_1}} \\ \hat{\mathbf{z}}_{\mathbf{i_2}} \\ \hat{\mathbf{z}}_{\mathbf{i_3}} \end{bmatrix} = \begin{bmatrix} -L_1 \mathbf{I} & \mathbf{I} & \mathbf{0} \\ -L_2 \mathbf{I} & \mathbf{0} & \mathbf{I} \\ -L_3 \mathbf{I} & \mathbf{0} & \mathbf{0} \end{bmatrix} \begin{bmatrix} \hat{\mathbf{z}}_{\mathbf{i_1}} \\ \hat{\mathbf{z}}_{\mathbf{i_2}} \\ \hat{\mathbf{z}}_{\mathbf{i_3}} \end{bmatrix} + \begin{bmatrix} L_1 \mathbf{A_i} & L_1 \mathbf{I} & -L_2 \mathbf{I} \\ L_2 \mathbf{A_i} & L_2 \mathbf{I} & -L_3 \mathbf{I} \\ L_3 \mathbf{A_i} & L_3 \mathbf{I} & \mathbf{0} \end{bmatrix} \begin{bmatrix} \mathbf{x_i} \\ L_1 \mathbf{x_i} \\ \mathbf{x_i} \end{bmatrix} + \begin{bmatrix} L_1 \mathbf{b_i} \\ L_2 \mathbf{b_i} \\ L_3 \mathbf{b_i} \end{bmatrix} u_i \tag{11}$$

By subtraction Eq. (11) from the time derivative of Eq. (9), the following disturbance estimation dynamics can be easily obtained.

$$\frac{d}{dt} \begin{bmatrix} \mathbf{z}_{\mathbf{i_1}} - \hat{\mathbf{z}}_{\mathbf{i_1}} \\ \mathbf{z}_{\mathbf{i_2}} - \hat{\mathbf{z}}_{\mathbf{i_2}} \\ \mathbf{z}_{\mathbf{i_3}} - \hat{\mathbf{z}}_{\mathbf{i_3}} \end{bmatrix} = \begin{bmatrix} -L_1 \mathbf{I} & \mathbf{I} & \mathbf{0} \\ -L_2 \mathbf{I} & \mathbf{0} & \mathbf{I} \\ -L_3 \mathbf{I} & \mathbf{0} & \mathbf{0} \end{bmatrix} \begin{bmatrix} \mathbf{z}_{\mathbf{i_1}} - \hat{\mathbf{z}}_{\mathbf{i_1}} \\ \mathbf{z}_{\mathbf{i_2}} - \hat{\mathbf{z}}_{\mathbf{i_2}} \\ \mathbf{z}_{\mathbf{i_3}} - \hat{\mathbf{z}}_{\mathbf{i_3}} \end{bmatrix} + \begin{bmatrix} \mathbf{0} \\ \mathbf{0} \\ \mathbf{I} \end{bmatrix} \dddot{\boldsymbol{\tau}}_{\mathbf{dis_i}} \tag{12}$$

Equation (12) shows that the second-order DOb is Bounded-Input Bounded-Output (BIBO) stable when $\|\dddot{\boldsymbol{\tau}}_{\mathbf{dis_i}}\| \leq \lambda_4$ and asymptotically stable when $\dddot{\boldsymbol{\tau}}_{\mathbf{dis_i}} = \mathbf{0}$.

### B. DOb-based Robust Motion Controller in State Space

Since the dynamic model of the compliant robot derived in Eq. (7) is observable, we can generate trajectory references of the robust motion controller using Brunovsky canonical form as proposed in [28]. To this end, let us obtain the canonical form of the dynamic model using Eq. (13)

$$\dot{\boldsymbol{\xi}}_\mathbf{i} = \boldsymbol{\Lambda}_\mathbf{i}\boldsymbol{\xi}_\mathbf{i} + \boldsymbol{\beta}_\mathbf{i}\nu - \boldsymbol{\Gamma}_\mathbf{i} \quad (13)$$

where $\boldsymbol{\xi}_\mathbf{i} = \mathbf{T}\mathbf{x}_\mathbf{i} = \begin{bmatrix} \xi_{i_1} & \cdots & \xi_{i_4} \end{bmatrix}^T \in \mathbb{R}^4$ ; $\boldsymbol{\Lambda}_\mathbf{i} = \mathbf{T}\mathbf{A}_\mathbf{i}\mathbf{T}^{-1} = \begin{bmatrix} \mathbf{0} & \mathbf{I} \\ & \mathbf{a}_{i_c}^\mathbf{T} \end{bmatrix} \in \mathbb{R}^{4\times 4}$ ; $\boldsymbol{\beta}_\mathbf{i} = \mathbf{T}\mathbf{b}_\mathbf{i} = \begin{bmatrix} \mathbf{0} \\ 1 \end{bmatrix} \in \mathbb{R}^4$ ; $\boldsymbol{\Gamma}_\mathbf{i} = \mathbf{T}\boldsymbol{\tau}_{\mathbf{dis}_\mathbf{i}} = \begin{bmatrix} \Gamma_{i_1} & \cdots & \Gamma_{i_4} \end{bmatrix}^T \in \mathbb{R}^4$ in which $\mathbf{T} \in \mathbb{R}^{4\times 4}$ is the transformation matrix of the Brunovsky canonical form [29].

The state vector of the canonical form can be directly derived from Eq. (13) as follows.

$$\boldsymbol{\xi}_\mathbf{i} = \begin{bmatrix} \xi_{i_1} \\ \xi_{i_2} \\ \xi_{i_3} \\ \xi_{i_4} \end{bmatrix} = \begin{bmatrix} \xi_{i_1} \\ \dot{\xi}_{i_1} + \Gamma_1 \\ \dot{\xi}_{i_2} + \Gamma_2 \\ \dot{\xi}_{i_3} + \Gamma_3 \end{bmatrix} = \begin{bmatrix} \xi_{i_1} \\ \dot{\xi}_{i_1} + \Gamma_1 \\ \ddot{\xi}_{i_1} + \dot{\Gamma}_1 + \Gamma_2 \\ \dddot{\xi}_{i_1} + \ddot{\Gamma}_1 + \dot{\Gamma}_2 + \Gamma_3 \end{bmatrix} \quad (14)$$

Hence, the state and control input references can be generated using Eq. (15) and Eq. (16).

$$\boldsymbol{\xi}_\mathbf{i}^\mathbf{ref} = \begin{bmatrix} \xi_{i_1}^{ref} \\ \xi_{i_2}^{ref} \\ \xi_{i_3}^{ref} \\ \xi_{i_4}^{ref} \end{bmatrix} = \begin{bmatrix} \xi_{i_1}^{ref} \\ \dot{\xi}_{i_1}^{ref} + \hat{\Gamma}_1 \\ \dot{\xi}_{i_2}^{ref} + \hat{\Gamma}_2 \\ \dot{\xi}_{i_3}^{ref} + \hat{\Gamma}_3 \end{bmatrix} = \begin{bmatrix} \xi_{i_1}^{ref} \\ \dot{\xi}_{i_1}^{ref} + \hat{\Gamma}_1 \\ \ddot{\xi}_{i_1}^{ref} + \hat{\dot{\Gamma}}_1 + \hat{\Gamma}_2 \\ \dddot{\xi}_{i_1}^{ref} + \hat{\ddot{\Gamma}}_1 + \hat{\dot{\Gamma}}_2 + \hat{\Gamma}_3 \end{bmatrix} \quad (15)$$

$$\nu = \ddddot{\xi}_{i_1}^{ref} + \mathbf{K}_\mathbf{i}\left(\boldsymbol{\xi}_\mathbf{i}^\mathbf{ref} - \boldsymbol{\xi}_\mathbf{i}\right) + \sum_{j=1}^{4} \hat{\Gamma}_{i_j}^{(4-j)} - \mathbf{a}_\mathbf{c}^\mathbf{T}\boldsymbol{\xi}_\mathbf{i}^\mathbf{ref} \quad (16)$$

where $\mathbf{K}_\mathbf{i} \in \mathbb{R}^{1\times 4}$ represents the control gain vector of the state space controller.

Equations (15) and (16) show that we need to estimate the disturbance vector and its successive first and second order derivatives to generate the state and control input references. Therefore, we need to use a higher-order DOb in the proposed robust motion controller synthesis.

### C. Stability of the Robust Motion Controller

The stability of the second order DOb, which is proven in the previous section, is a necessary condition in the robust motion controller synthesis. However, we still need to prove the stability of the overall robust motion controller.

To this end, let us use the following Lyapunov function candidate.

$$\mathbf{V} = \boldsymbol{\xi}_\mathbf{i}^T \mathbf{P}_\mathbf{i} \boldsymbol{\xi}_\mathbf{i} \quad (17)$$

$$\left(\boldsymbol{\Lambda}_\mathbf{i} - \boldsymbol{\beta}_\mathbf{i}\mathbf{K}_\mathbf{i}\right)^T \mathbf{P}_\mathbf{i} + \mathbf{P}_\mathbf{i}\left(\boldsymbol{\Lambda}_\mathbf{i} - \boldsymbol{\beta}_\mathbf{i}\mathbf{K}_\mathbf{i}\right) = -\mathbf{Q}_\mathbf{i} \quad (18)$$

where $\mathbf{P}$ and $\mathbf{Q}$ are positive definite matrices that we use to tune the control gain vector of the state pace controller $\mathbf{K}_\mathbf{i}$.

The derivate of the Lyapunov function candidate is derived using Eq. (13) and Eq. (17) as follows.

$$\dot{\mathbf{V}} = -\boldsymbol{\xi}_\mathbf{i}^T \mathbf{Q}_\mathbf{i}\boldsymbol{\xi}_\mathbf{i} + 2\boldsymbol{\xi}_\mathbf{i}^T \mathbf{P}_\mathbf{i}\left(\hat{\boldsymbol{\Gamma}}_\mathbf{i} - \boldsymbol{\Gamma}_\mathbf{i}\right) \quad (19)$$

Equation (19) shows that all states of the system are uniformly ultimately bounded when the disturbance estimation error is bounded, i.e., the second order DOb is BIBO stable. When the second order DOb is asymptotically stable, the derivative of the Lyapunov function candidate is negative definite. In other words, asymptotic stability is obtained for the robust motion controller.

This analysis shows that the proposed robust motion controller can suppress the matched and mismatched disturbances, including the disturbances of SEAs, external loads, and the interactive forces of a multi-body system.

## IV. SIMULATIONS

This section verifies the proposed robust position and force controllers via simulations of a redundant compliant robot manipulator illustrated in Fig. 2. It is assumed that the 3-DOF planar robot is built using SEAs at each joint. While the robot manipulator can freely move within its workspace in position control, it physically interacts with a silver-ball that constraints the movement of the manipulator in force control. Virtual Reality Toolbox of MATLAB/SIMULINK is employed to develop a dynamic simulator for the compliant planar robot arm.

Let us first consider the robust position control problem of the compliant robot manipulator. It is assumed that a 1kg object is placed between second and third joints and a 2.5kg object is placed at tip after 3 seconds. It is also assumed that the dynamic model of the robot manipulator cannot be precisely identified, i.e., inertia $\mathbf{m}(\mathbf{q_m})$ and Coriolis/centripetal $\mathbf{c}(\mathbf{q_m}, \dot{\mathbf{q}}_\mathbf{m})$ matrices include uncertainties and there are unknown disturbances such as friction and hysteresis at joints. Step and sinusoidal position references are applied to each joint to verify the performance of the robust position controller in regulation and trajectory tracking control applications. Figures 3 and 4 show that the proposed robust position controller can precisely track the step and sinusoidal trajectory references when the robot manipulator suffers from internal and external disturbances.

Let us now consider the robust force control problem of the compliant robot manipulator. Similarly, it is assumed that the dynamic model of the robot manipulator cannot be precisely identified. However, external loads are not applied to the robot manipulator in force control. Figures 5 and 6 show force control results when the end point of the planar robot physically interacts with a soft and a stiff silver-ball. It is clear from these figures that the proposed robust force controller allows us to conduct high-performance force control applications when environmental impedance changes.

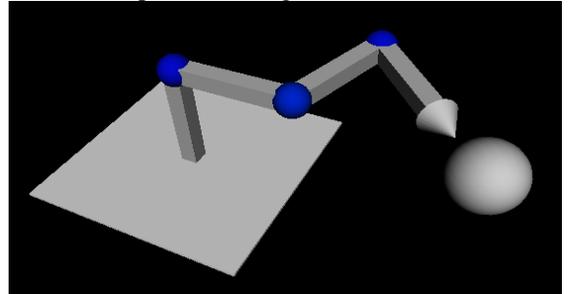

Fig. 2: Redundant planar robot manipulator.

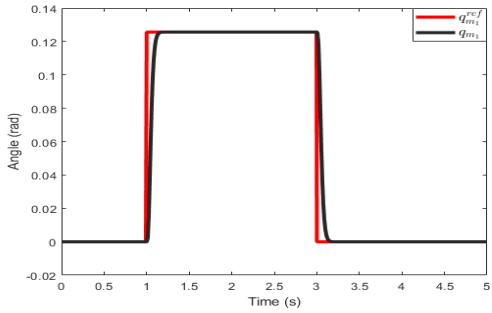

a) Regulation control at joint 1.

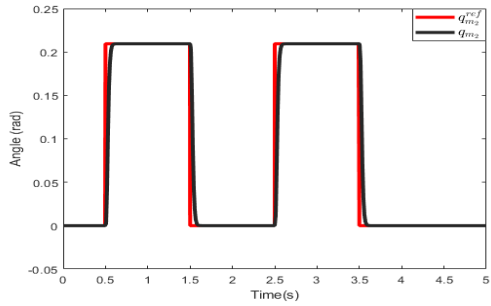

b) Regulation control at joint 2.

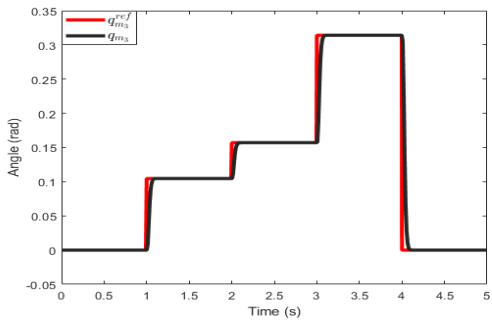

c) Regulation control at joint 3.

Fig. 3: Robust Position (Regulation) Control.

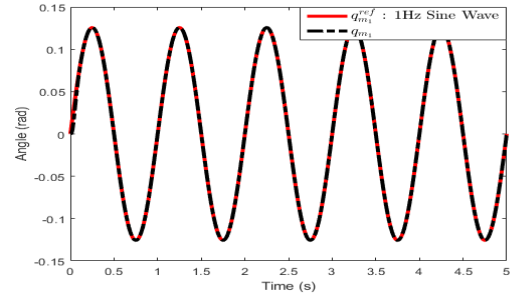

a) Trajectory tracking control at joint 1.

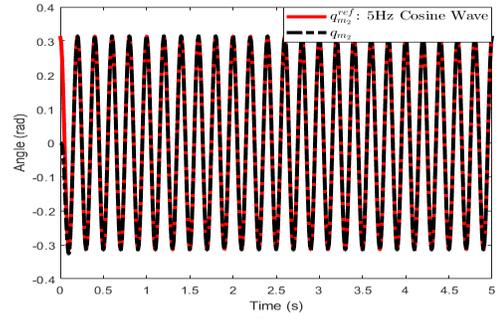

b) Trajectory tracking control at joint 2.

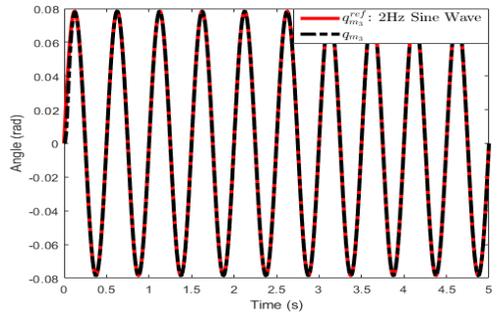

c) Trajectory tracking control at joint 3.

Fig. 4: Robust Position (Trajectory Tracking) Control.

## V. CONCLUSIONS

In this paper, a new decentralised robust motion controller is proposed for compliant robots driven by SEAs. This robust motion controller enables one to conduct high-performance position and force control tasks by precisely suppressing internal and external disturbance using a second-order DOb and a novel controller design method in state space. High-performance of the robust motion controller is verified by simulations for both position and force control tasks. The promising simulation results motivate us to conduct an experimental study in the future.

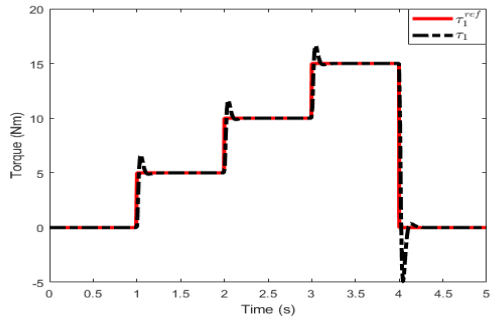

a) Regulation control at joint 1.

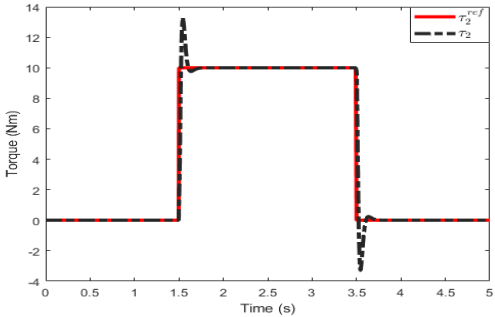

b) Regulation control at joint 2.

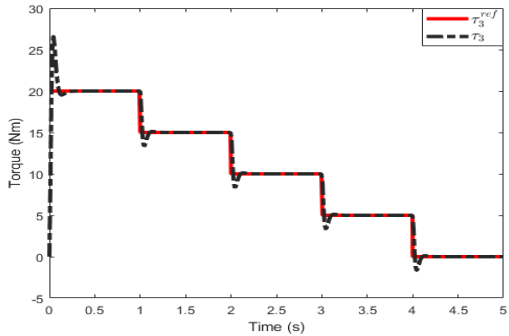

c) Regulation control at joint 3.

Fig. 5: Robust Force Control when the robot physically interacts with a soft environment.

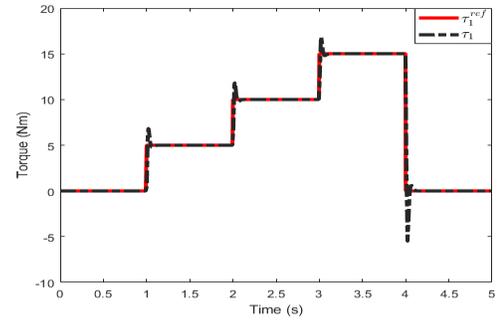

a) Regulation control at joint 1.

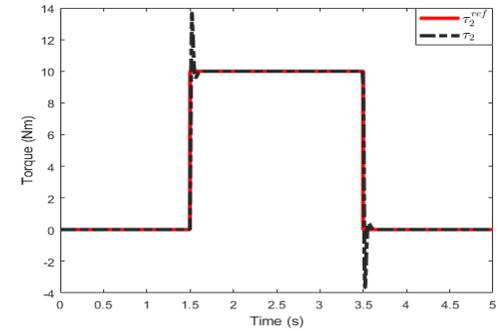

b) Regulation control at joint 2.

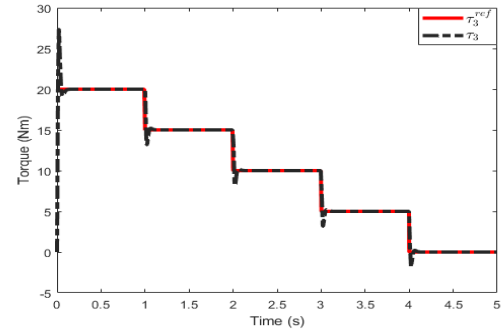

c) Regulation control at joint 3.

Fig. 6: Robust Force Control when the robot physically interacts with a stiff environment.